\documentclass[conference]{IEEEtran}
\IEEEoverridecommandlockouts
\usepackage{cite}
\usepackage{amsmath,amssymb,amsfonts}
\usepackage{algorithmic}
\usepackage{graphicx}
\usepackage{textcomp}
\usepackage{xcolor}
\ifCLASSOPTIONcompsoc
\usepackage[caption=false,font=normalsize,labelfont=sf,textfont=sf]{subfig}
\else
\usepackage[caption=false,font=footnotesize]{subfig}
\fi

\def\BibTeX{{\rm B\kern-.05em{\sc i\kern-.025em b}\kern-.08em
    T\kern-.1667em\lower.7ex\hbox{E}\kern-.125emX}}
\begin{document}

\title{LoGAN: Generating Logos with a Generative Adversarial Neural Network Conditioned on color}

\author{\IEEEauthorblockN{Ajkel Mino}
\IEEEauthorblockA{
\textit{Department of Data Science}
\textit{and Knowledge Engineering}\\
\textit{Maastricht University}\\
Maastricht, Netherlands \\
ajkimino@gmail.com}
\and
\IEEEauthorblockN{Gerasimos Spanakis}
\IEEEauthorblockA{
\textit{Department of Data Science}
\textit{and Knowledge Engineering}\\
\textit{Maastricht University}\\
Maastricht, Netherlands\\
jerry.spanakis@maastrichtuniversity.nl}
\thanks{We gratefully acknowledge the support of Mediaan in this research especially for providing the necessary computing resources.}
}

\maketitle

\begin{abstract}
Designing a logo is a long, complicated, and expensive process for any designer. However, recent advancements in generative algorithms provide models that could offer a possible solution. Logos are multi-modal, have very few categorical properties, and do not have a continuous latent space. Yet, conditional generative adversarial networks can be used to generate logos that could help designers in their creative process. We propose LoGAN: an improved auxiliary classifier Wasserstein generative adversarial neural network (with gradient penalty) that is able to generate logos conditioned on twelve different colors. In 768 generated instances (12 classes and 64 logos per class), when looking at the most prominent color, the conditional generation part of the model has an overall precision and recall of 0.8 and 0.7 respectively. LoGAN's results offer a first glance at how artificial intelligence can be used to assist designers in their creative process and open promising future directions, such as including more descriptive labels which will provide a more exhaustive and easy-to-use system. 
\end{abstract}


\section{Introduction}

Designing a logo is a lengthy process that requires continuous collaboration between the designers and their clients. Each drafted logo takes time and effort to be designed, which turns this creative process into a tedious and expensive endeavor. 

Recent advancements in generative models suggest a possible use of artificial intelligence as a solution to this problem. Specifically, Generative Adversarial Neural Networks (GANs) \cite{gan-ian}, which learn how to mimic any distribution of data. They consist of two neural networks, a generator and a discriminator, that are contended against each other, trying to reach a Nash Equilibrium \cite{nash-equilibrium} in a minimax game. 

Whilst GANs and other generative models have been used to generate almost anything, from MNIST digits \cite{gan-ian} to anime faces \cite{anime-gan}, Mario levels \cite{mario-gan} and fake celebrities \cite{progressivegrowing}, logo generation has yet to receive thorough exploration. One possible explanation is that logos do not contain a hierarchy of nested segments, which networks can learn and try to reproduce. Furthermore, their latent space is not continuous, meaning that not every generated logo will necessarily be aesthetically pleasing. To the best of our knowledge, Sage, et al. \cite{logopaper} to be only one to have tackled this problem thus far. They propose a clustered approach for dealing with multi-modal data, specifically logos. Logos get assigned synthetic labels, defined by the cluster they belong to, and a GAN is trained, conditioned on these labels. However, synthetic labels do not provide enough flexibility. The need for more descriptive labels arises: labels that could better convey what is in the logos, and present designers with more detailed selection criteria.
 
This paper provides a first step to more expressive and informative labels instead of synthetic ones, which is achieved by defining logos using their most prominent color. Twelve color classes are introduced: black, blue, brown, cyan, gray, green, orange, pink, purple, red, white, and yellow. We propose LoGAN: an improved Auxiliary Classifier Wasserstein Generative Adversarial Neural Network (with Gradient Penalty) (AC-WGAN-GP) that is able to generate logos conditioned on the aforementioned twelve colors. 

The rest of this paper is structured as follows: In Section 2 the related work and background will be discussed, then in Section 3 the proposed architecture be conveyed, followed by Section 4, where the dataset and labeling process will be explained. Section 5 presents the experimental results, consisting of the training details and results, and finally we conclude and discuss some possible extensions of this work in Section 6.

\section{Background \& Related Work}

First proposed in 2014 by Goodfellow et al. \cite{gan-ian}, GANs have seen a rise in popularity during the last couple of years, after possible solutions to training instability and mode collapse were introduced \cite{convergence-and-stability, convergence, improved-gan}. 

\subsection{Generative Adversarial Networks}
\begin{figure*}[ht]
	\centering
	\includegraphics[width=\textwidth, height = 0.30\textheight]{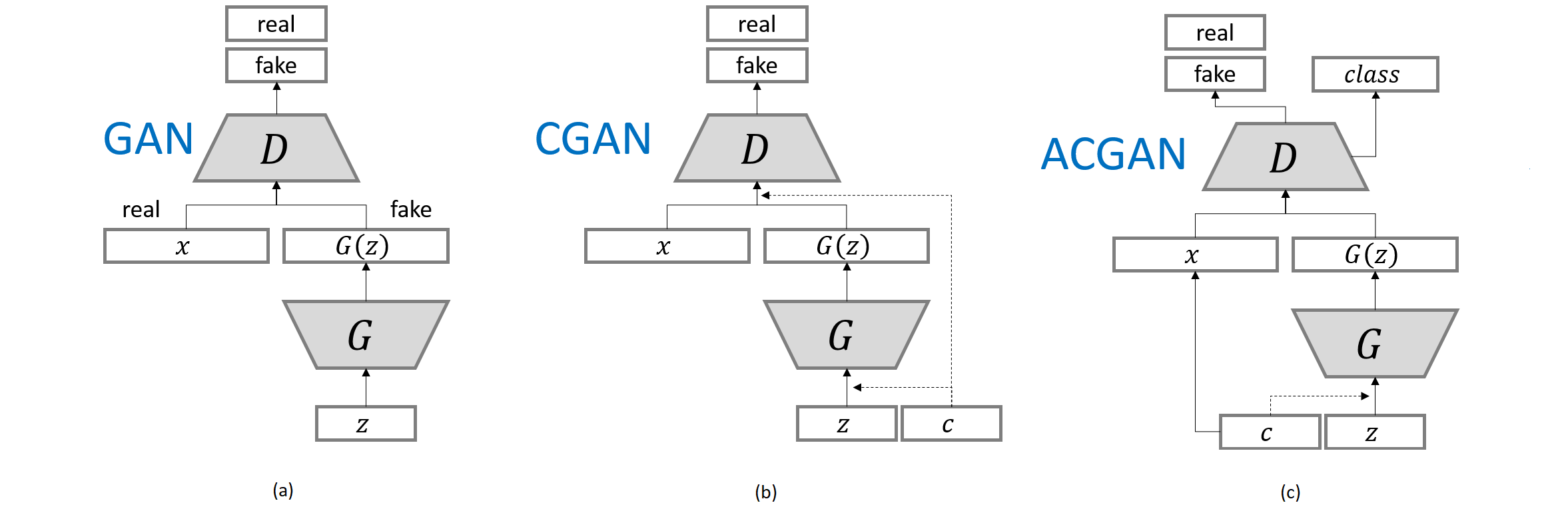}
    \caption{GAN, conditional GAN (CGAN) and auxiliary classifier GAN (ACGAN) architectures, where $x$ denotes the real image, $c$ the class label, $z$ the noise vector, $G$ the Generator, and $D$ the Discriminator.}
    \label{fig:gan-architectures}
\end{figure*}

GANs (architecture depicted in \figurename \ref{fig:gan-architectures}(a)) consist of two different neural networks, a generator and a discriminator, that are trained simultaneously in a competitive manner. 

The generator is fed a noise vector ($G(z)$) from a probability distribution ($p_z$), and outputs a generated data-point (fake image). The discriminator takes its input either from the generator (the fake image) ($D(G(z))$) or from the training set (the real image) ($D(x)$) and is trained to distinguish between the two. The discriminator and the generator play a two-player minimax game with value function \eqref{eq:minmiaxgan}, where the discriminator tries to maximize V, while the generator tries to minimize it. 

\begin{align}
	\underset{G} \min \hspace{1mm} \underset{D} \max \hspace{1mm} V(D,G) =   & \!\begin{aligned}[t] \mathbb{E}_{x \mathtt{\sim} p_{data}}[\log D(x)] &\\ 
     + \mathbb{E}_{z \mathtt{\sim} p_{z}}[\log (1-D(G(z)))] \end{aligned}
    \label{eq:minmiaxgan}
\end{align}

\subsubsection{Objective functions}
While GANs are able to generate high-quality images, they are notoriously difficult to train. They suffer from problems like training stability, non-convergence and mode collapse. Multiple improvements have been suggested to fix these problems \cite{convergence-and-stability, convergence, improved-gan}; including using deep convolutional layers for the networks \cite{dcgan} and modified objective functions i.e. to least-squares \cite{LSGAN} or to Wasserstein distance between the distributions \cite{wgan, began, wgan-gp}.

\subsection{Conditionality}
Conditional generation with GANs entails using labeled data to generate images based on a certain class. The two types that will be discussed in the subsections below are Conditional GANs and Auxiliary Classifier GANs.

\subsubsection{Conditional GANs}
In a conditional GAN (CGAN) \cite{cgan} (architecture depicted in \figurename \ref{fig:gan-architectures}(b)) the discriminator and the generator are conditioned on $c$, which could be a class label or some data from another modality. The input and $c$ are combined in a joint hidden representation and fed as an additional input layer in both networks.

\subsubsection{Auxiliary Classifier GANs}
Contrary to CGANs, in Auxiliary Classifier GANs (AC-GAN) \cite{acgan} (architecture depicted in \figurename \ref{fig:gan-architectures}(c)) the latent space $z$ is conditioned on the class label. The discriminator is forced to identify fake and real images, as well as the class of the image, irrespective of whether it is fake or real.

\subsection{GAN Applications}
These different GAN architectures have been used for numerous purposes, including (but not limited to): higher-quality image generation \cite{progressivegrowing, stackgan}, image blending \cite{gp}, image super-resolution \cite{superrez} and object detection \cite{objectdetect}.

Up until the writing of this paper, logo generation has previously only been investigated by Sage et al. \cite{logopaper}, who accomplish three main things:
\begin{itemize}
\item Define the Large Logo Dataset (LLD) \cite{data}
\item Synthetically label the logos by clustering them using the ResNet Classifier network
\item Build a GAN conditioned on these synthetic labels  
\end{itemize}

However, as these labels are defined from computer-generated clusters, they do not necessarily provide intuitive classes for a human designer. More descriptive labels, that could provide designers with more detailed selection criteria and better convey what is in the logos are needed.

This paper offers a solution to the previous problem by:
\begin{enumerate}
\item Using twelve colors to define the logo classes
\item Defining LoGAN: An AC-WGAN-GP that can conditionally generate logos
\end{enumerate} 

\section{Proposed Architecture}

\begin{figure}[h]
	\includegraphics[width=7cm, height=0.30\textheight]{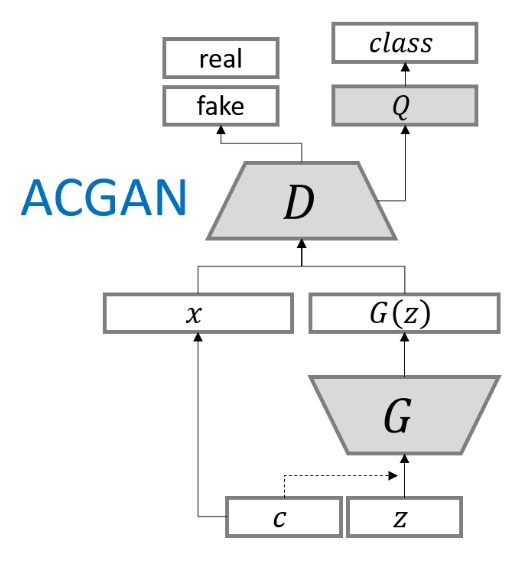}
    \caption{LoGAN architecture, where $x$ denotes the real image, $c$ the class label, $z$ the noise vector, $G$ the Generator, $D$ the Discriminator and $Q$ the Classifier}
    \label{fig:logan_architecture}
\end{figure}

The proposed architecture for LoGAN is an Auxiliary Classifier Wasserstein Generative Adversarial Neural Network with gradient penalty (AC-WGAN-GP), depicted on \figurename \ref{fig:logan_architecture}. LoGAN is based on the ACGAN architecture 
, with the main difference being that it consists of three neural networks, namely the discriminator, the generator and an additional classification network
\footnote{Code for this paper is available via  \texttt{https://github.com/ajki/LoGAN}}. 
The latter is responsible for assisting the discriminator in classifying the logos, as the original classification loss from AC-GAN was dominated by the Wasserstein distance used by the WGAN-GP.

In an original AC-GAN \cite{acgan} the discriminator is forced to optimize the classification loss together with the real-fake loss. This is shown in equations \eqref{eq:acgan_loss_correct_source} \& \eqref{eq:acgan_loss_correct_class}, which describe the loss from defining the source of the image (training set or generator) and the loss from defining the class of the image respectively. The adversarial aspect of AC-GAN comes as the discriminator tries to maximize $\mathcal{L}^{ACGAN}_{Class} + \mathcal{L}^{ACGAN}_{Source}$, whilst the generator tries to maximize $\mathcal{L}^{ACGAN}_{Class} - \mathcal{L}^{ACGAN}_{Source}$ \cite{acgan}.

\begin{align}
	 \mathcal{L}^{ACGAN}_{Source} = & \!\begin{aligned}[t] & \mathbb{E}[\log {P(S = real |X_{real})}] \\
  & + \mathbb{E}[\log {P(S=fake|X_{fake})}]  \end{aligned}
    \label{eq:acgan_loss_correct_source}
\end{align}

\begin{equation}
    \mathcal{L}^{ACGAN}_{Class} = \mathbb{E}[\log {P(C = c |X_{real})}] + \mathbb{E}[\log {P(C=c|X_{fake})}]
    \label{eq:acgan_loss_correct_class}
\end{equation}

Since WGAN-GP is more stable while training, the proposed architecture makes use of the WGAN-GP loss, instead of the AC-GAN loss. The loss function for the discriminator and generator of LoGAN are the same as a WGAN-GP \cite{wgan-gp}, stated in equations \eqref{eq:wgangp_loss_d} \& \eqref{eq:wgangp_loss_g}.

\begin{align}
    \mathcal{L}^{WGANGP}_{D} =  & \!\begin{aligned}[t] & - \mathbb{E}_{x \mathtt{\sim} p_{d}}[D(x)] + \mathbb{E}_{\hat{x} \mathtt{\sim} p_{g}}[D(\hat{x})] \\ 
    & + \lambda \mathbb{E}_{\hat{x} \mathtt{\sim} p_{g}} [(||\nabla D(\alpha x + (1 - \alpha \hat{x}))||_{2} - 1)^{2}] \end{aligned}
    \label{eq:wgangp_loss_d}
\end{align}

\begin{equation}
    \mathcal{L}^{WGANGP}_{G} = - \mathbb{E}_{\hat{x} \mathtt{\sim} p_{g}}[D(\hat{x})]
    \label{eq:wgangp_loss_g}
\end{equation}


The loss of the additional classifier network, is defined as:
\begin{equation}
    \mathcal{L}^{LoGAN}_{Q} = \mathbb{E}_{x,y}[\log {Q(y|x)}]
    \label{eq:wgangp_loss_c}
\end{equation}

To avoid instability during training and mode collapse certain measures were taken. The generator and classifier were trained once for every 5 iterations of discriminator training, as suggested by Gulrajan et al. \cite{wgan-gp}, $z$ was sampled from a Gaussian Distribution \cite{sampling}, and batch normalization was used \cite{batchnorm}.

\section{Data}
The dataset used to train LoGAN is the LLD-icons dataset \cite{data}, which consists of  486'377 $32 \times 32$ icons.

\subsubsection{Labeling} 
\begin{figure}[h]				
	\includegraphics[width=0.5\textwidth]{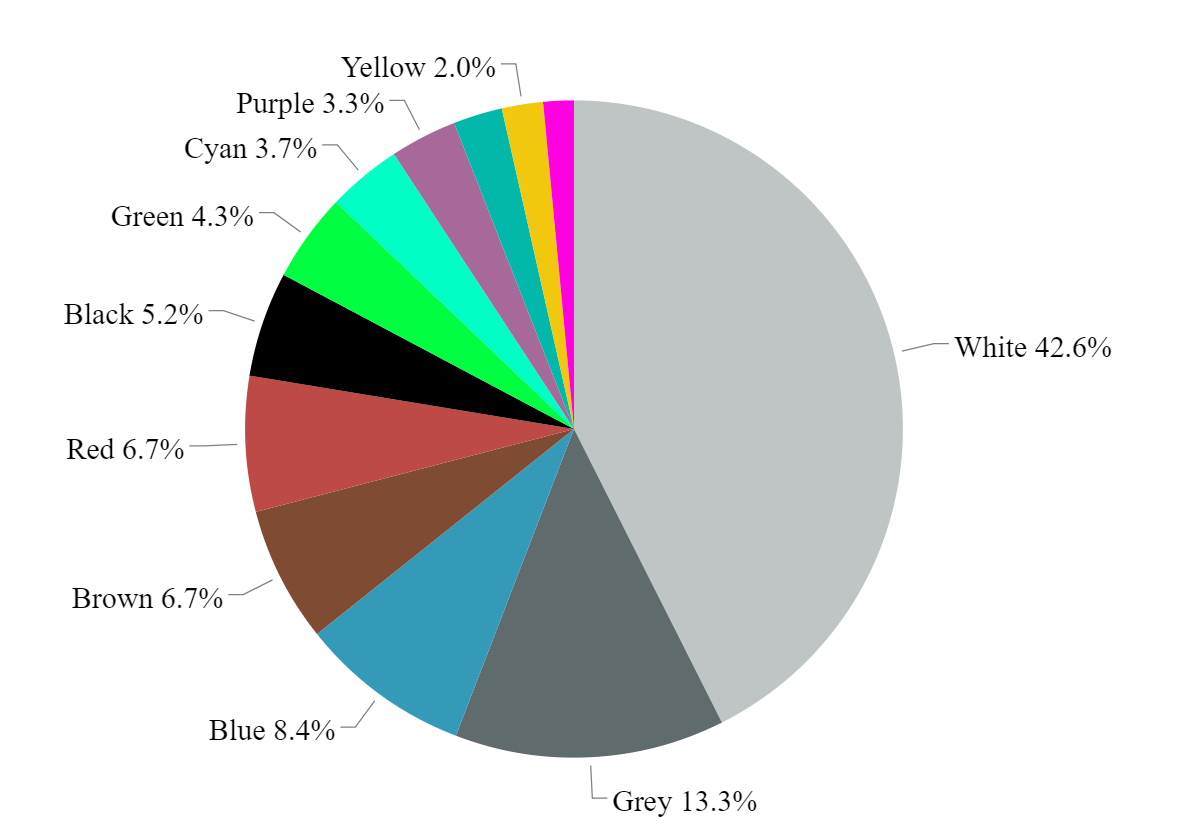}
    \caption{Dataset distribution by class. }
    \label{fig:classdistribution}
\end{figure}

To extract the most prominent color from the image a k-Means algorithm with $k = 3$ was used to define the RGB values of the centroids. The algorithm makes use of the MiniBatchKMeans implementation from sci-kit learn package, and the package webcolors was used to turn the RGB values into color words. The preliminary colors consisted of X11 color names \footnote{A list of the X11 color names, and the class grouping of RGB values can be found on Wikipedia at : https://en.wikipedia.org/wiki/Web\_colors\#X11\_color\_names}, which were grouped into 12 main classes: black, blue, brown, cyan, gray, green, orange, pink, purple, red, white, and yellow.
The class distribution can be observed in \figurename \ref{fig:classdistribution}.

\section{Experimental Results}
In this section, the training process, quality evaluation and the results obtained from the model \footnote{The model was trained on a Windows 10 machine, with a Tesla K80 GPU, for 400 epochs, which lasted around three full days.} will be presented. 

\subsection{Training}
\begin{figure}[!t]
\centering
\subfloat[]{\includegraphics[width=3.5in]{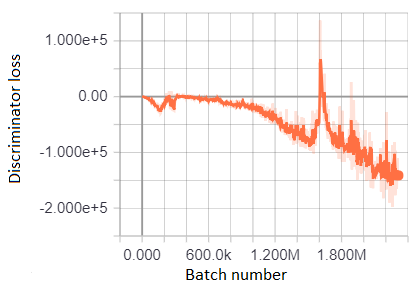}
\label{subfig:discriminator_loss}}
\hfil
\subfloat[]{\includegraphics[width=3.5in]{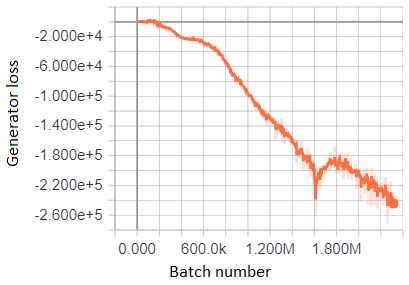}
\label{subfig:generator_loss}}
\hfil
\subfloat[]{\includegraphics[width=3.5in]{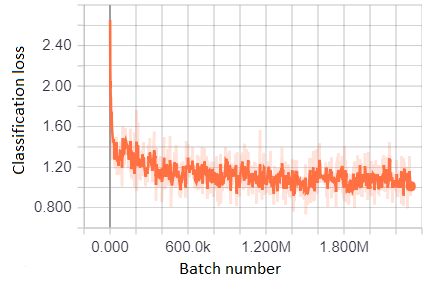}
\label{subfig:classifier_loss}}

\caption{Discriminator, Generator and Classifier losses, where the X-axis represents the batch number, and the Y-axis denotes the loss value for that certain batch.}
 	\label{fig:generator_discriminator_loss}
\end{figure}

\figurename \ref{fig:generator_discriminator_loss} shows the loss graphs per batch for the discriminator \ref{subfig:discriminator_loss}, generator \ref{subfig:generator_loss} and classifier \ref{subfig:classifier_loss}. It can be observed that both the discriminator and the generator have not converged as the loss graphs have a downward trend. This does, however, not imply improper training as neither WGAN nor WGAN-GP are guaranteed to reach convergence \cite{convergence}. 

\begin{figure}
	\centering
	\includegraphics[width=3.5in]{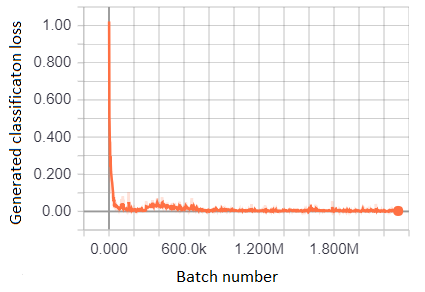}
    \caption{Classification loss for generated images.}
    \label{fig:classification_loss_f}
\end{figure}

The classifier on the other hand has converged with a loss value close to 1. Further investigation into the classification loss shows that the classification loss for fake images has converged to zero (loss depicted on \figurename \ref{fig:classification_loss_f}). This means that the generated images get classified correctly. 

\subsection{Quality evaluation}
The network is expected to generate logo-resembling images that have clear color definition. In each epoch, 64 logos will be generated per class. The top three most prominent colors in the logos will be extracted, and the generated pairs and triplets will be analyzed. For each class ($c$), the precision \eqref{eq:precision}, recall \eqref{eq:recall} and F1-score \eqref{eq:f1} will be measured for the most prominent color in the logo.

\begin{equation}
	Precision = \frac{correctly\_generated\_as(c)}{total\_generated\_as(c)}
	\label{eq:precision}
\end{equation}

\begin{equation}
	Recall = \frac{correctly\_generated\_as(c)}{total\_actual(c)}
	\label{eq:recall}
\end{equation}

\begin{equation}
	F1 = 2 * \frac{Precision * Recall}{Precision + Recall}
	\label{eq:f1}
\end{equation}

\subsection{Results}

\begin{figure*}[h]
	\includegraphics[width=\textwidth]{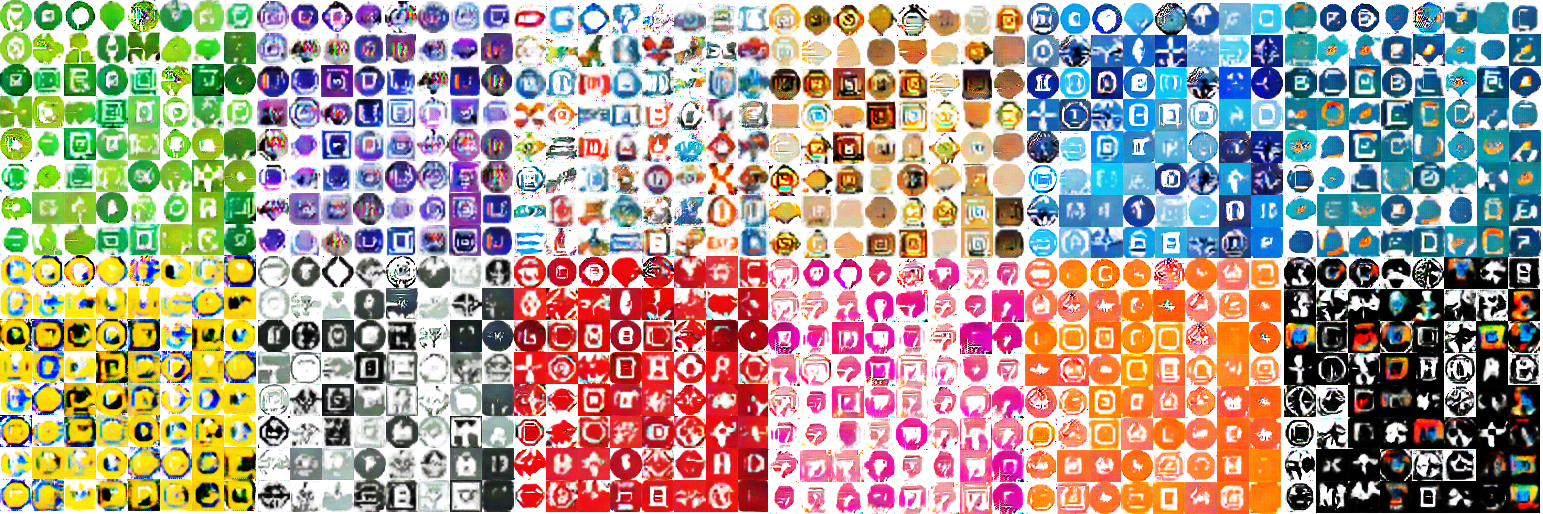}
    \caption{Results from the generation of 64 logos per class after 400 epochs of training. Classes from left to right top to bottom: green, purple, white, brown, blue, cyan, yellow, gray, red, pink, orange, black. }
    \label{fig:class-generation}
\end{figure*}

The results for the class conditioned generation after 400 epochs are shown in \figurename \ref{fig:class-generation}. As expected, a slight blurriness can be noticed on the generated logos. This is because the training images are only $32\times32$ pixels. Despite the blurriness, the generated logos resemble feasible ones. The generated images are mainly dominated by round and square shapes. Even irregular shapes have been generated, for example the heart and the x in the array of white logos. A look-alike of the Google Chrome logo is also present in the midst of the cyan class.

The precision, recall and F1-score of the class conditioned generation after 400 epochs is shown on Table \ref{tab:400epochresults}. The precision scores are relatively high with the exception of white and gray. This is because the most predominantly white logos are small logos with a lot of white or transparent space around. Consequently many small logos generated in other colors are going to be classified as white. Similarly, because of its neutrality the gray class also appears a lot in other classes.

\begin{table}[h]
\centering
\caption{Precision, Recall and F1-score of the most prominent color in each of the logos in \figurename \ref{fig:class-generation}.}
\label{tab:400epochresults}
\begin{tabular}{c|ccc}
Class   & Precision & Recall   & F1\\ \hline
black   & 0,95  & 0,86 	& 0,90 \\
blue    & 0,73  & 0,69  & 0,71 \\
brown   & 0,63  & 0,47  & 0,55 \\
cyan    & 0,98  & 0,66  & 0,79 \\
gray    & 0,57  & 0,50  & 0,53 \\
green   & 1     & 0,80 	& 0,89 \\
orange  & 0,96  & 0,80 	& 0,87 \\
pink    & 0,95  & 0,30 	& 0,45 \\
purple  & 0,65  & 0,41  & 0,50 \\
red     & 0,84  & 0,92 	& 0,88 \\
white   & 0,24  & 0,83 	& 0,38 \\
yellow  & 0,96  & 0,78  & 0,86 \\ \hline
Average & 0,79  & 0,67 	& 0,69
\end{tabular}
\end{table}

The recall values are lower than the precision overall, with red having the highest recall at 0.92 and pink having the lowest at 0.3. Most of the pink labeled logos generated belong to class white. 
 
The F1-score marks black on top with 0.90, and white on the bottom with 0.38. Based on Equation \eqref{eq:f1}, and the fact that white has the lowest precision, its F1-score is also expected to be low.

\begin{figure*}[h]				
	\includegraphics[width=\textwidth, height = .4\textheight]{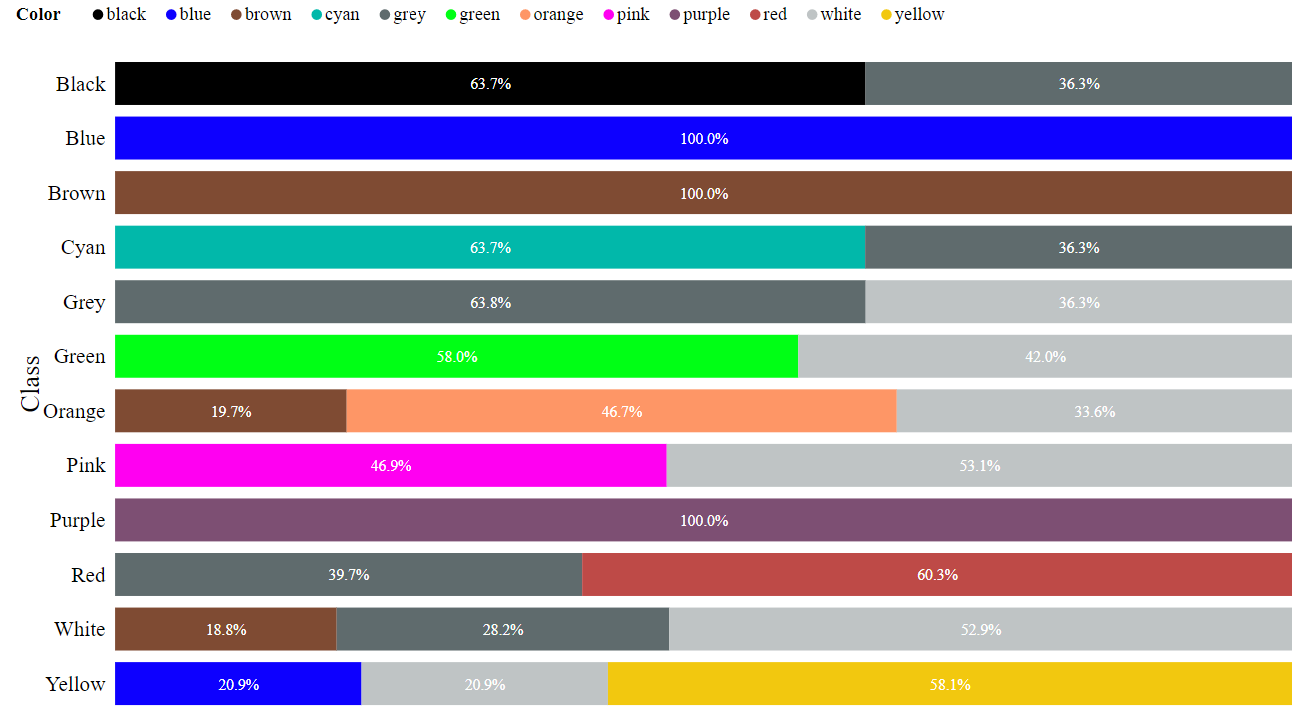}
    \caption{Distribution of the top three generated colors per class.}
    \label{fig:distribution}
\end{figure*}

\figurename \ref{fig:distribution} shows the distribution of the top three generated colors in each class. This distribution is calculated by extracting the top three colors from each logo, and distinguishing the most prominent ones in each category. As expected from the results in Table \ref{tab:400epochresults}, white and gray are present in the top three for many of the classes. Some interesting combinations include the orange class, where the color brown appears, and the yellow class, where the color blue appears. There are three classes which get generated using only shades from their own class (blue, brown, purple). However, if we consider black and white as shades of gray, that number rises to five.

\section{Conclusion \& Future work}
In this paper, it was shown how designers can be assisted in their creative process for logo design by modern artificial intelligence techniques, namely Generative Adversarial Networks. The proposed model can successfully create logos if given a certain keyword, which in our case consisted of the most prominent color in the logo. This class of keywords can be considered descriptive as it provides a property of the logo that is easy for humans to distinguish.

The proposed architecture consists of an Auxilliary Classification Wasserstein GAN with gradient penalty (AC-WGAN-GP) and generates logos conditioned on twelve colors. This helps designers in their creative process and while brainstorming, making logo design cheaper and less time-consuming. As the generated logos have very low resolution, they can serve as a very rough first draft of a final logo, or as a means of inspiration for the designer. Regarding the results, the classifier converged, and the generated logos meet the requirements of the class they belong to. This is backed up by precision, recall, and comparison with other logos.
 
Despite the promising results, a limitation of the approach is the blurriness of the generated logos. At the same time, color is not a stand-alone keyword for defining a logo. Higher resolution training images and other labels such as the shape of the logo or the focus of the company would provide valuable input, thus improving the results. 

Possible extensions to this work include conditioning on more labels, such as the shape of the logo. However, as logos do not always have clear geometrical shapes, possible issues could include logos with text only or logos with irregular shapes. These issues could be overcome by splitting the dataset into two main groups, logos with an obvious geometrical shape, e.g. quadrilateral, circular or triangular; and logos with a non-regular shape, e.g. text, letters, hearts, etc. Extracting shapes from the first group can be easily done using packages like \texttt{OpenCV}. The non-regular group of shapes may be a bit more challenging to label. A possible path is to use a tool such as \texttt{tessaract-ocr} perform optical character recognition to extract the text in the logo, and use it as the label. As for the irregularly shaped logos, those could be put in an extra class with a label such as 'others'.

Finally, another possible set of labels for the logos could be gathered for the LLD-logos dataset (which also provides higher quality images) by gathering the most used words to describe the company the logo belongs to. Combining these labels with word embedding models could potentially introduce a semantic meaning to the logos, further boosting the interpretability of the current approach.

\bibliography{references}

\begin{thebibliography}{10}
\providecommand{\url}[1]{#1}
\csname url@samestyle\endcsname
\providecommand{\newblock}{\relax}
\providecommand{\bibinfo}[2]{#2}
\providecommand{\BIBentrySTDinterwordspacing}{\spaceskip=0pt\relax}
\providecommand{\BIBentryALTinterwordstretchfactor}{4}
\providecommand{\BIBentryALTinterwordspacing}{\spaceskip=\fontdimen2\font plus
\BIBentryALTinterwordstretchfactor\fontdimen3\font minus
  \fontdimen4\font\relax}
\providecommand{\BIBforeignlanguage}[2]{{%
\expandafter\ifx\csname l@#1\endcsname\relax
\typeout{** WARNING: IEEEtran.bst: No hyphenation pattern has been}%
\typeout{** loaded for the language `#1'. Using the pattern for}%
\typeout{** the default language instead.}%
\else
\language=\csname l@#1\endcsname
\fi
#2}}
\providecommand{\BIBdecl}{\relax}
\BIBdecl

\bibitem{gan-ian}
I.~Goodfellow, J.~Pouget-Abadie, M.~Mirza, B.~Xu, D.~Warde-Farley, S.~Ozair,
  A.~Courville, and Y.~Bengio, ``Generative adversarial nets,'' in
  \emph{Advances in neural information processing systems}, 2014, pp.
  2672--2680.

\bibitem{nash-equilibrium}
J.~F. Nash \emph{et~al.}, ``Equilibrium points in n-person games,''
  \emph{Proceedings of the national academy of sciences}, vol.~36, no.~1, pp.
  48--49, 1950.

\bibitem{anime-gan}
Y.~Jin, J.~Zhang, M.~Li, Y.~Tian, H.~Zhu, and Z.~Fang, ``Towards the automatic
  anime characters creation with generative adversarial networks,''
  unpublished.

\bibitem{mario-gan}
\BIBentryALTinterwordspacing
V.~Volz, J.~Schrum, J.~Liu, S.~M. Lucas, A.~M. Smith, and S.~Risi, ``Evolving
  mario levels in the latent space of a deep convolutional generative
  adversarial network,'' in \emph{Proceedings of the Genetic and Evolutionary
  Computation Conference (GECCO 2018)}.\hskip 1em plus 0.5em minus 0.4em\relax
  New York, NY, USA: ACM, July 2018. [Online]. Available:
  \url{http://doi.acm.org/10.1145/3205455.3205517}
\BIBentrySTDinterwordspacing

\bibitem{progressivegrowing}
T.~Karras, T.~Aila, S.~Laine, and J.~Lehtinen, ``Progressive growing of gans
  for improved quality, stability, and variation,'' unpublished.

\bibitem{logopaper}
A.~Sage, E.~Agustsson, R.~Timofte, and L.~Van~Gool, ``Logo synthesis and
  manipulation with clustered generative adversarial networks,'' unpublished.

\bibitem{convergence-and-stability}
N.~Kodali, J.~Abernethy, J.~Hays, and Z.~Kira, ``On convergence and stability
  of gans,'' unpublished.

\bibitem{convergence}
L.~Mescheder, A.~Geiger, and S.~Nowozin, ``Which training methods for gans do
  actually converge?” arxiv preprint,'' unpublished.

\bibitem{improved-gan}
T.~Salimans, I.~Goodfellow, W.~Zaremba, V.~Cheung, A.~Radford, and X.~Chen,
  ``Improved techniques for training gans,'' in \emph{Advances in Neural
  Information Processing Systems}, 2016, pp. 2234--2242.

\bibitem{dcgan}
A.~Radford, L.~Metz, and S.~Chintala, ``Unsupervised representation learning
  with deep convolutional generative adversarial networks,'' unpublished.

\bibitem{LSGAN}
X.~Mao, Q.~Li, H.~Xie, R.~Y. Lau, Z.~Wang, and S.~P. Smolley, ``Least squares
  generative adversarial networks,'' in \emph{2017 IEEE International
  Conference on Computer Vision (ICCV)}.\hskip 1em plus 0.5em minus 0.4em\relax
  IEEE, 2017, pp. 2813--2821.

\bibitem{wgan}
M.~Arjovsky, S.~Chintala, and L.~Bottou, ``Wasserstein gan,'' unpublished.

\bibitem{began}
D.~Berthelot, T.~Schumm, and L.~Metz, ``Began: Boundary equilibrium generative
  adversarial networks,'' unpublished.

\bibitem{wgan-gp}
I.~Gulrajani, F.~Ahmed, M.~Arjovsky, V.~Dumoulin, and A.~C. Courville,
  ``Improved training of wasserstein gans,'' in \emph{Advances in Neural
  Information Processing Systems}, 2017, pp. 5769--5779.

\bibitem{cgan}
M.~Mirza and S.~Osindero, ``Conditional generative adversarial nets,''
  unpublished.

\bibitem{acgan}
A.~Odena, C.~Olah, and J.~Shlens, ``Conditional image synthesis with auxiliary
  classifier gans,'' unpublished.

\bibitem{stackgan}
H.~Zhang, T.~Xu, H.~Li, S.~Zhang, X.~Huang, X.~Wang, and D.~Metaxas,
  ``Stackgan: Text to photo-realistic image synthesis with stacked generative
  adversarial networks,'' in \emph{IEEE Int. Conf. Comput. Vision (ICCV)},
  2017, pp. 5907--5915.

\bibitem{gp}
H.~Wu, S.~Zheng, J.~Zhang, and K.~Huang, ``Gp-gan: Towards realistic
  high-resolution image blending,'' unpublished.

\bibitem{superrez}
C.~Ledig, L.~Theis, F.~Husz{\'a}r, J.~Caballero, A.~Cunningham, A.~Acosta,
  A.~Aitken, A.~Tejani, J.~Totz, Z.~Wang \emph{et~al.}, ``Photo-realistic
  single image super-resolution using a generative adversarial network,''
  unpublished.

\bibitem{objectdetect}
J.~Li, X.~Liang, Y.~Wei, T.~Xu, J.~Feng, and S.~Yan, ``Perceptual generative
  adversarial networks for small object detection,'' in \emph{IEEE CVPR}, 2017.

\bibitem{data}
A.~Sage, E.~Agustsson, R.~Timofte, and L.~Van~Gool, ``Lld - large logo dataset
  - version 0.1,'' \url{https://data.vision.ee.ethz.ch/cvl/lld}, 2017.

\bibitem{sampling}
T.~White, ``Sampling generative networks: Notes on a few effective
  techniques,'' unpublished.

\bibitem{batchnorm}
S.~Ioffe and C.~Szegedy, ``Batch normalization: Accelerating deep network
  training by reducing internal covariate shift,'' unpublished.

\end{thebibliography}

\end{document}